# Welldefined Decision Scenarios

**Thomas D. Nielsen**　　　**Finn V. Jensen**
Department of Computer Science
Aalborg University
Fredrik Bayers Vej 7C, DK-9220, Aalborg Øst, Denmark

## Abstract

Influence diagrams serve as a powerful tool for modelling symmetric decision problems. When solving an influence diagram we determine a set of strategies for the decisions involved. A strategy for a decision variable is in principle a function over its past. However, some of the past may be irrelevant for the decision, and for computational reasons it is important not to deal with redundant variables in the strategies. We show that current methods (e.g. the *Decision Bayes-ball* algorithm [Shachter, 1998]) do not determine the relevant past, and we present a complete algorithm.

Actually, this paper takes a more general outset: When formulating a decision scenario as an influence diagram, a linear temporal ordering of the decisions variables is required. This constraint ensures that the decision scenario is welldefined. However, the structure of a decision scenario often yields certain decisions conditionally independent, and it is therefore unnecessary to impose a linear temporal ordering on the decisions. In this paper we deal with partial influence diagrams i.e. influence diagrams with only a partial temporal ordering specified. We present a set of conditions which are necessary and sufficient to ensure that a partial influence diagram is welldefined. These conditions are used as a basis for the construction of an algorithm for determining whether or not a partial influence diagram is welldefined.

## 1 INTRODUCTION

Graphical modelling for decision support systems is getting more and more widespread. First of all, graphical modelling is an appealing way to think of and communicate on the underlying structure of the domain in question, but it also helps the modeller to focus on structure rather than calculations. *Influence diagrams* (ID) serve as a powerful modelling tool for symmetric decision problems with several decisions. However, IDs require a linear temporal ordering of the decisions, and this is often felt as an unnecessary constraint. E.g. if no information is gathered between two decisions, then these decisions can be taken independently of each other. This type of very obvious temporal independence can be handled by a computer system, but temporal independence may be more complicated. For example, observing a variable $A$ immediately before taking a decision $D$ need not have any impact on this particular decision. Hence, we look for an operational characterization of temporal independence in IDs.

The advantages of having an operational characterization of temporal independence are twofold. To take the most obvious advantage first. When a computer system solves an ID it basicly eliminates the variables in reverse temporal order (see [Shachter, 1986], [Shenoy, 1992], [Jensen et al., 1994] and [Zhang, 1998]); eliminating a variable produces a table (function) over all non-eliminated neighbours. However, the reverse temporal order of elimination has a tendency to create very large tables (usually much larger than for Bayesian networks of the same complexity). Thus, if we could relax the temporal order to a partial order, we would have more freedom when looking for a good elimination sequence. To say it another way. When a decision variable $D$ is eliminated we create a strategy for $D$ given its past. If we can reduce the past to contain only the variables required for taking that decision, we have reduced the domain for the strategy function.

The second advantage has to do with the modelling process. That is, will we allow two decisions to be taken independently of each other? Or, do we allow an observation to be made independently of a certain decision? Hence, we work with partial specifications of IDs, and would therefore like to know whether or not this partial structure is ambiguous, and if it is ambiguous we would like the system to give suggestions for further specification of the temporal order. An unambiguous partial influence diagram is said to represent a *welldefined decision scenario*.

In this paper we give a set of operational rules for determining whether or not a partial influence diagram represents a welldefined decision scenario. These rules are used as a basis for an algorithm for answering this question. The algorithm can furthermore be used in a dialogue between computer and user to pinpoint how to change the model in order to make it unambiguous.

In section 2 we formally introduce IDs, and the terms and notations used throughout this paper. In section 3 we define a partial ID as a generalization of the traditional ID, and we give a semantic as well as a syntactic characterization of conditions ensuring that a partial influence diagram is unambiguous.

## 2 INFLUENCE DIAGRAMS

An ID can be seen as a belief network augmented with decision variables and utility functions. Thus, the nodes in the ID can be partitioned into three disjoint subsets; *chance nodes*, *decision nodes* and *value nodes*.

The chance nodes (drawn as circles) correspond to *chance variables*, and represent events which are not under the direct control of the decision maker. The decision nodes (drawn as squares) correspond to *decision variables* and represent actions under the direct control of the decision maker. In the remainder of this section we assume a total ordering of the decision nodes, indicating the order in which the decisions are made.[1] Furthermore, we will use the concept of node and variable interchangeably if this does not introduce any inconsistency, and we assume that no *barren nodes* are specified by the ID since they have no impact on the decisions.[2]

The set of value nodes (drawn as diamonds) defines a set of *utility functions*, indicating the local utility for a given configuration of the variables in their domain. The total utility is the sum or the product of the local utilities; in the remainder of this paper we assume that the total utility is the sum of the local utilities.

---

[1]The ordering of the decision nodes is traditionally represented by a directed path which includes all decision nodes.

[2]A chance node or a decision node is said to be barren if it does not precede any other node, or if all its descendants are barren.

With each chance variable and decision variable $X$ we associate a *state space* $W_X$ which denotes the set of possible outcomes/decision alternatives for $X$. For a set $\mathcal{U}'$ of variables we define the state space as $W_{\mathcal{U}'} = \times \{W_X | X \in \mathcal{U}'\}$.

The uncertainty associated with each chance variable $A$ is represented by a *conditional probability function* $P(A|\mathcal{P}_A) : W_{A \cup \mathcal{P}_A} \to [0;1]$, where $\mathcal{P}_A$ denotes the immediate predecessors of $A$.

The arcs in an ID can be partitioned into three disjoint subsets, corresponding to the type of node they go into. Arcs into value nodes represent functional dependencies by indicating the domain of the associated utility function. Arcs into chance nodes, denoted *dependency arcs*, represent probabilistic dependencies, whereas arcs into decision nodes, denoted *informational arcs*, imply information precedence; if there is an arc from a node $X$ to a decision node $D$ then the state of $X$ is known when decision $D$ is made.

Let $\mathcal{U}_C$ be the set of chance variables and let $\mathcal{U}_D = \{D_1, D_2, \ldots, D_n\}$ be the set of decision variables. Assuming that the decision variables are ordered by index, the set of informational arcs induces a partitioning of $U_C$ into a collection of disjoint subsets $C_0, C_1, \ldots, C_n$. The set $C_j$ denotes the chance variables observed between decision $D_j$ and $D_{j+1}$ thus, the variables in $C_j$ occur as immediate predecessors of $D_{j+1}$. This induces a *partial order* $\prec$ on $\mathcal{U} = \mathcal{U}_C \cup \mathcal{U}_D$ i.e. $C_0 \prec D_1 \prec C_1 \prec \cdots \prec D_n \prec C_n$

The set of variables known to the decision maker when deciding on $D_j$ is called the *informational predecessors* of $D_j$ and is denoted $\mathrm{pred}(D_j)$. Assuming "no-forgetting" the set $\mathrm{pred}(D_j)$ corresponds to the set of variables that occur before $D_j$ under $\prec$. Moreover, based on the "no-forgetting" assumption we can assume that an ID does not specify any redundant no-forgetting arcs i.e. a chance node can be an immediate predecessor of at most one decision node.

### 2.1 EVALUATION

When evaluating an ID we identify a strategy for the decision variables; a strategy can be seen as a prescription of responses to earlier observations and decisions. The evaluation is usually performed according to the *maximum expected utility principle*, which states that we should always choose an alternative that maximizes the expected utility.

**Definition 1.** Let $\mathcal{I}$ be an ID and let $\mathcal{U}_D$ denote the decision variables in $\mathcal{I}$. A *strategy* is a set of functions $\Delta = \{\delta_D | D \in \mathcal{U}_D\}$, where $\delta_D$ is a *decision function* given by:

$$\delta_D : W_{\mathrm{pred}(D)} \to W_D,$$

A strategy that maximizes the expected utility is termed an *optimal strategy*.

In general, the optimal strategy for a decision variable $D_k$ in an ID $\mathcal{I}$ is given by:

$$\delta_{D_k}(C_0, D_1, \dots, C_{k-1}) =$$
$$\arg\max_{D_k} \sum_{C_k} P(C_k | C_0, D_1, \dots, C_{k-1}, D_k) \rho_{D_{k+1}} \quad (1)$$

where $\rho_{D_{k+1}}$ is the maximum expected utility function for decision $D_{k+1}$:

$$\rho_{D_{k+1}}(C_0, D_1, \dots, C_k) =$$
$$\max_{D_{k+1}} \sum_{C_{k+1}} P(C_{k+1} | C_0, D_1, \dots, C_k, D_{k+1}) \rho_{D_{k+2}}$$

By continuously expanding Equation 1, we get the following expression for the optimal strategy for $D_k$:

$$\delta_{D_k}(C_0, D_1, \dots, C_{k-1}) = \arg\max_{D_k} \sum_{C_k} \cdots \max_{D_n} \sum_{C_n}$$
$$P(C_k, \dots, C_n | C_0, D_1, \dots, C_{k-1}, D_1, \dots, D_n)(\psi_1 + \cdots + \psi_l),$$

where $\psi_1, \dots \psi_l$ are the utility functions specified by $\mathcal{I}$.

The expression above conveys that the variables are to be eliminated w.r.t. an elimination sequence which is consistent with the partial order, and in what follows we define a *legal elimination sequence* as a bijection $\alpha : \mathcal{U} \leftrightarrow \{1, 2, \dots, |\mathcal{U}|\}$, where $X \prec Y$ implies $\alpha(X) < \alpha(Y)$. Note that a legal elimination sequence is not necessarily unique, since the chance variables in the sets $C_j$ can be commuted. Even so, any two legal elimination sequences result in the same optimal strategy since the decision variables are ordered totally and '$\sum$' operations commute; the total ordering of the decision variables ensures that the relative elimination order for any pair of variables of opposite type is invariant under the legal elimination sequences (this is needed since a 'max' operation and a '$\sum$' operation do not commute in general).

## 3   REPRESENTING DECISION PROBLEMS UNAMBIGUOUSLY

In the section above we described the evaluation of an ID w.r.t. the maximum expected utility principle. The underlying assumption was a total ordering of the decision variables ensuring that the optimal strategy is independent of the legal elimination sequences.

However, it is in general not necessary to have a total ordering of the decision variables (if $C_k = \emptyset$ then $D_k$ and $D_{k+1}$ can be commuted). This can also be seen from the optimal strategy for a decision variable (equation 1), where the elimination order for any two adjacent variables of opposite type may be permuted if the variables do not occur in the same function.

### 3.1   PARTIAL INFLUENCE DIAGRAMS AND DECISION SCENARIOS

Given that a total ordering of the decision variables may not be needed, we define a *partial influence diagram* (PID) as a directed acyclic graph consisting of decision nodes, chance nodes and value nodes, assuming that value nodes have no children. Notice, that chance nodes may have several decision nodes as immediate successors, and that no ordering is imposed on the decisions. Additionally, we define a *realization* of a PID as an attachment of functions to the appropriate variables i.e. the chance variables are associated with conditional probability functions and the value nodes are associated with utility functions.

Since the semantics of a PID correspond to the semantics of an ID, a PID induces a partial order $\prec$ on the nodes $\mathcal{U}_C \cup \mathcal{U}_D$, as defined by the transitive closure of the following relation (see figure 1):

- $Y \prec D_i$, if $(Y, D_i)$ is a directed arc in $\mathcal{I}$ ($D_i \in \mathcal{U}_D$).

- $D_i \prec Y$, if $(D_i, X_1, X_2, \dots, X_m, Y)$ is a directed path in $\mathcal{I}$ ($Y \in \mathcal{U}_C \cup \mathcal{U}_D$ and $D_i \in \mathcal{U}_D$).

- $D_i \prec A$, if $A \not\prec D_j$ for all $D_j \in \mathcal{U}_D$ ($A \in \mathcal{U}_C$ and $D_i \in \mathcal{U}_D$).

- $D_i \prec A$, if $A \not\prec D_i$ and $\exists D_j \in \mathcal{U}_D$ s.t. $D_i \prec D_j$ and $A \prec D_j$ ($A \in \mathcal{U}_C$ and $D_i \in \mathcal{U}_D$).

In what follows we say that two different nodes $X$ and $Y$ in a PID $\mathcal{I}$ are *incompatible* if $X \not\prec Y$ and $Y \not\prec X$. Note that a chance node $A$ is incompatible with a decision node $D$ if there exists a decision node $D'$ s.t. $D$ and $D'$ are incompatible and $(A, D')$ is an informational arc in $\mathcal{I}$ and $A \not\in \text{pred}(D)$(see figure 1).

As for the traditional ID we seek to identify an optimal strategy when evaluating a PID. Since the optimal strategy for a decision variable may be dependent on variables observed, we define a total order $<$ for a PID.

**Definition 2.** Let $\mathcal{I}$ be a PID and let $\mathcal{U} = \mathcal{U}_D \cup \mathcal{U}_C$ denote the set of decision variables and chance variables contained in $\mathcal{I}$. A *total ordering* of $\mathcal{I}$ is a bijection $\beta : \mathcal{U} \leftrightarrow \{1, 2, \dots, |\mathcal{U}|\}$. A total ordering of $\mathcal{I}$ is said to be an *admissible total ordering* if $X \prec Y$ implies that $\beta(X) < \beta(Y)$, where $\prec$ is the partial order induced by $\mathcal{I}$.

In what follows $<_\beta$ will denote the total ordering $\beta$ s.t. $X <_\beta Y$ if $\beta(X) < \beta(Y)$ (the index $\beta$ will be omitted if this does not introduce any confusion). E.g. $E < F < D_2 < B < D_1 < G < D_4 < D_3 < C_4$ is a total ordering of the PID in figure 1, but it is not admissible since it contradicts $D_1 \prec D_2$.

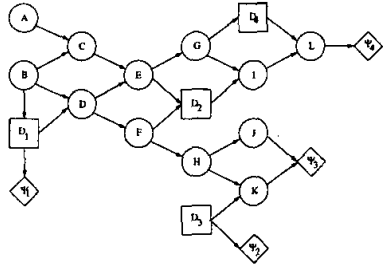

Figure 1: The figure represents a PID which specifies the partial order $\{B\} \prec D_1 \prec \{E, F, G, D_2, D_4\} \prec C_4$, $\{B\} \prec D_1 \prec \{E, F\} \prec D_2 \prec C_4$, $\{B\} \prec D_1 \prec \{G\} \prec D_4 \prec C_4$ and $D_3 \prec C_4$ ($C_4$ denotes the chance variables observed (possibly never) after deciding on all the decisions). Thus, $D_2$, $D_3$ and $D_4$ are pairwise incompatible, whereas $D_1$ and $D_4$ are not. Furthermore, it can be seen that $F$ is incompatible with $D_4$.

Notice, that an admissible total ordering of a PID $\mathcal{I}$ implies that $\mathcal{I}$ can be seen as an ID (assuming that redundant no-forgetting arcs have been removed).

Based on the informational predecessors for a decision variable, we define a *strategy relative to a total order* $<$, as a set of functions $\Delta^< = \{\delta_D^< | D \in \mathcal{U}_D\}$, where $\delta_D^<$ is a *decision function* given by:

$$\delta_D^< : W_{\text{pred}(D)^<} \to W_D,$$

where $\text{pred}(D)^< = \{X | X < D\}$ (the index $<$ in $\text{pred}(D)^<$ will be omitted if this does not introduce any confusion). Given a realization of a PID $\mathcal{I}$, we term a strategy relative to $<$, an *optimal strategy relative to* $<$ if the strategy maximizes the expected utility. Likewise we term a decision function $\delta_D^<$ contained in an optimal strategy relative to $<$, an *optimal strategy for $D$ relative to* $<$. Note that an optimal strategy for a decision variable $D$ relative to $<$ does not necessarily depend on all the variables observed. Hence we say that an observed variable $X$ is *required* for $D$ w.r.t. $<$ if there is a realization of $\mathcal{I}$ s.t. the optimal strategy for $D$ relative to $<$ is dependent on the state of $X$.

Since a total order for a PID need not be admissible, we define an *admissible optimal strategy* for a realization of a PID as an optimal strategy relative to an admissible total order.

**Definition 3.** A realization of a PID $\mathcal{I}$ is said to *define a decision scenario* if all admissible optimal strategies for $\mathcal{I}$ are identical. A PID is said to *define a decision scenario* if all its realizations define a decision scenario.

The above definition characterizes the class of PIDs which can be considered welldefined, since the set of admissible total orderings for an PID $\mathcal{I}$ can be seen as the legal elimination sequences for $\mathcal{I}$ (Note that the traditional ID defines a decision scenario). Moreover, in correspondence with the permutations of chance variables in any legal elimination sequence for an ID, we define the following relation for any admissible total order.

**Definition 4.** Let $<$ be an admissible total order, and let $X$ and $Y$ be two neighbouring variables under $<$, fulfilling one of the following three conditions:

- $X$ and $Y$ are both chance variables.

- $X$ and $Y$ are both decision variables.

- $X$ and $Y$ are incompatible.

The ordering $<'$ obtained from $<$ by permuting $X$ and $Y$ according to the rules above is said to be *C-equivalent* with $<$, denoted $\equiv_C <'$.

**Proposition 1.** *The transitive closure of $\equiv_C$ is an equivalence relation.*

**Theorem 1.** *All admissible orderings of a partial order $\prec$ are C-equivalent.*

*Proof.* It is sufficient to prove the following claim: Let $<$ be an admissible total ordering, and let $X$ and $Y$ be incompatible s.t. $X < Y$. Then the ordering obtained from $<$ by permuting $X$ and $Y$ is C-equivalent with $<$.

Assume the claim not to be true. Then there exists an admissible total ordering $<$ and a pair of incompatible variables $X, Y$ which can not be permuted. Let $X$ and $Y$ be such that the segment between $X$ and $Y$ under $<$ is minimal. That is, it is not possible to find any other admissible total order with an incompatible non-permutable pair of variables closer than $X$ and $Y$ under $<$:

$$X < X_1 < \cdots < X_n < Y$$

Now, start with $X$ and follow $<$ until we reach an incompatible variable $X_i$; we know that at least when we reach $Y$ we will meet an incompatible variable. If $X_i = Y$ then $Y$ and $X_{i-1}$ can be permuted, and we have an admissible ordering with an incompatible non-permutable pair closer than the closest. If $i \leq n$ then $X$ and $X_i$ are incompatible. If they can not be permuted we have a pair of incompatible non-permutable variables closer than the closest. If they can be permuted we also obtain a closer pair. $\square$

So, we are looking for a set of necessary and sufficient conditions ensuring that all admissible orderings yield the same set of strategies. Actually, we will look for conditions ensuring that orderings, C-equivalent with an admissible ordering, yield the same strategies; this is a bit broader as we allow permutation of two neighbouring decision variables. From Theorem 1 we infer that we can narrow down the scope to neighbouring variables of opposite type (in general neighbouring variables of opposite type can not be permuted without affecting the strategies). Hence, we look for a necessary and sufficient set of conditions granting commutation of two incompatible neighbouring variables of opposite type.

**Definition 5.** Let $\mathcal{I}$ be a PID and let $A$ be a chance variable incompatible with a decision variable $D$ in $\mathcal{I}$. Then $A$ is said to be *significant* for $D$ if there is a realization and an admissible total order $<$ for $\mathcal{I}$ s.t.

- $A$ occurs immediately before $D$ under $<$.

- The optimal strategy for $D$ relative to $<$ is different from the one achieved by permuting $A$ and $D$ in $<$.

A chance variable is said to be *significant for D relative to* $<$ if the above conditions are satisfied w.r.t. $<$.

Notice that if a chance variable $A$ is significant for $D$ w.r.t. $<$ then $A$ is required for $D$ w.r.t. $<$.

Based on the above definitions we present the following theorem which characterizes the constraints necessary and sufficient for a PID to define a decision scenario.

**Theorem 2.** The PID $\mathcal{I}$ defines a decision scenario if and only if for each decision variable $D$ there does not exist a chance variable $A$ significant for $D$.

*Proof.* Follows immediately from Theorem 1 and Definition 5. □

So, we have reduced the task to the following: Let $\mathcal{I}$ be a PID, and let $A$ be a chance variable incompatible with a decision variable $D$. Is $A$ significant for $D$?

[Shachter, 1998] presents an algorithm for determining the so called requisite information for a decision variable in an ID. Unfortunately, the algorithm does not meet our needs as shown by the following example.

**Example 1.** When running the algorithm *Decision Bayes-ball*[Shachter, 1998] on the ID depicted in figure 2, the chance variable $B$ is marked as requisite for decision $D_1$. However, $B$ is not relevant for the optimal strategy for $D_1$, i.e. the elimination order of $B$ relative to $D_1$ is of no importance when considering the optimal strategy for $D_1$. □

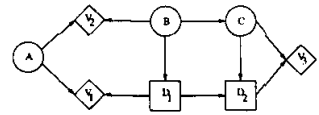

Figure 2: The algorithm Decision Bayes-ball marks the chance variable $B$ as requisite for $D_1$.

The following method, which corresponds to iteratively replacing decisions by their strategies, has the same drawback. For an ID we start off with the *moral graph* i.e. informational arcs are removed, undirected arcs are added between nodes with a common child and finally, value nodes are removed together with the directions on the arcs. When eliminating a decision variable $D$ the resulting set of neighbours $N(D)$ is a subset of pred($D$). This set of neighbours is invariant w.r.t. the legal elimination sequences, and it is characterized as the set of variables connected in the moral graph to $D$ through a path with no intermediate variable in pred($D$). As $N(D)$ contains all the information relevant for determining the optimal strategy for $D$, it is a candidate for the relevant past. However $N(D)$ may contain variables insignificant for $D$ as can be seen from the ID depicted in Figure 2; $B$ is contained in the neighbouring set for $D_1$ as the elimination of $A$ produces a fill-in between $B$ and $D_1$.

So, neither Decision Bayes-ball nor the elimination method presented above is fine-grained enough to detect all independencies. The problem is that $N(D)$ may contain variables relevant for the maximum expected utility for $D$; the maximum expected utility for $D$ may cover utility functions having no influence on the optimal strategy for $D$. This means that we need to characterize and identify the utility functions on which the optimal strategy for $D$ depends.

**Definition 6.** The utility function $\psi$ is *relevant* for $D$ w.r.t. the admissible total order $<$ for $\mathcal{I}$, if there exists two realizations $R_1$ and $R_2$ of $\mathcal{I}$ who only differ on $\psi$ s.t. the optimal strategies for $D$ relative to $<$ are different in $R_1$ and $R_2$.

We need to determine the structural constraints necessary and sufficient for a utility function to be relevant for a decision variable, and based on this characterization we shall define the constraints necessary and sufficient for a chance variable to be significant for a given decision variable.

## 3.2 RELEVANT UTILITY FUNCTIONS - EXAMPLES AND RULES

The optimal strategy for a decision variable $D$ is based on the assumption that we always adhere to the maximum expected utility principle. Hence, if deciding on $D$ can influence a future decision $D'$ then the utility functions relevant for $D'$ may be relevant for $D$ also. From this observation together with the expression for the optimal strategy for $D$ (see equation 1), we present the following metarules. For notational convenience we shall sometimes treat uninstantiated decision nodes as chance nodes with an even prior distribution. Moreover, since a utility function is termed relevant w.r.t. an admissible total order we will mainly consider IDs in the section.

**Metarule 1.** $\psi$ is relevant for $D$ in $\mathcal{I}$ if there is a realization of $\mathcal{I}$ s.t. $D$ has an impact on the expected utility for $\psi$.

**Metarule 2.** $\psi$ is relevant for $D$ in $\mathcal{I}$ if there is a realization of $\mathcal{I}$ and a future decision $D'$ s.t. $D$ has an impact on $D'$, for which $\psi$ is relevant.

**Metarule 3.** If none of the metarules above can be applied then $\psi$ is not relevant for $D$.

The following examples present a set of IDs where we identify the utility functions relevant for a given decision variable. The properties relating to these examples will be generalized to arbitrary IDs, which will serve as a basis for determining the structural constraints necessary for a utility function to be relevant for a given decision variable.

**Example 2.** Consider the ID depicted in figure 3 and assume that the conditional probability functions are specified s.t. the state of a variable corresponds to the state of its parent.

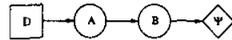

Figure 3: The figure represents an ID, where the utility function $\psi$ is relevant for the decision variable $D$.

It is easy to specify two realizations of $\psi$ s.t. the optimal strategies relative to those realizations differ i.e. $\psi$ is relevant for $D_1$.  $\square$

Now, assume an arbitrary ID $\mathcal{I}$ in which there exists a directed path from a decision node $D$ to a value node $\psi$ (excluding informational arcs), and assume a realization $R$ of $\mathcal{I}$. Since the conditional probability functions associated with the variables on the path from $D$ to $\psi$ can be specified s.t. deciding on $D$ has an impact on the expected utility for $\psi$, it follows that $\psi$ is relevant for $D$.

**Rule 1.** *Let $\mathcal{I}$ be a PID, and let $\bar{\mathcal{I}}$ denote $\mathcal{I}$ without informational arcs. The utility function $\psi$ is relevant for the decision variable $D$ if there exists a directed path from $D$ to $\psi$ in $\bar{\mathcal{I}}$.*

This rule is equivalent to Metarule 1, as can be seen from the mathematical expression corresponding to this metarule: $\psi$ is relevant for $D$ if $P(\text{dom}(\psi)|D, \text{pred}(D))$ is a function of $D$, where $\text{dom}(\psi)$ is the chance variables in the domain of $\psi$ (uninstantiated decision variables are treated as chance variables). The conditional probability function $P(\text{dom}(\psi)|\text{pred}(D), D)$ is a function of $D$ if $D$ is d-connected to a variable $A \in \text{dom}(\psi)$ given $\text{pred}(D)$. However, this implies that there exists a directed path from $D$ to $\psi$ in $\bar{\mathcal{I}}$. Conversely, if there exists a directed path from $D$ to $\psi$ in $\bar{\mathcal{I}}$, then $D$ is d-connected to a variable $A \in \text{dom}(\psi)$ given $\text{pred}(D)$.

Table 1: The utility function $\psi'(C, D_4)$.

| $\psi'(C, D_4)$ | $c_1$ | $c_2$ |
|---|---|---|
| $d_1$ | 10 | 0 |
| $d_2$ | 0 | 9 |

The following examples illustrate, that in order to identify all the utility functions relevant for a given decision variable $D$, it is in general not sufficient only to consider those utility functions to which there exist a directed path from $D$.

**Example 3.** When deciding on $D_4$ in the ID depicted in Figure 4 we want to maximize $\psi'$. Now, as the decision variable $D_2$ is d-connected to $C$ given $\text{pred}(D_4)$, it follows that the decision made w.r.t. $D_2$ may change our belief in $C$ (when deciding on $D_4$) and thereby influence $D_4$ through $\psi'$ ($D_2$ is required for $D_4$). Hence $\psi'$ is relevant for $D_2$, which is also true for $\psi$ as $D_2 \in \text{dom}(\psi)$. Moreover, knowledge of $A$ may likewise change our belief in $C$ when deciding on $D_4$, and since $A$ is influenced by $D_1$ it follows that $D_1$ has an impact on $D_2$ since knowledge of $D_1$ can be taken into account when deciding on $D_2$ ($D_1$ is required for $D_2$). Thus, $\psi'$ is relevant for both $D_1$ and $D_2$ conveying that $\psi$ is relevant for $D_1$ also.

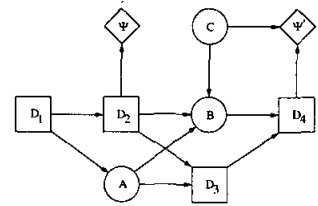

Figure 4: The figure represents an ID where both $\psi$ and $\psi'$ are relevant for $D_1$.

This can also be seen by considering the utility function specified by table 1 together with the functions $P(C) = (0, 5; 0, 5)$, $\psi_1(D_2) = (0; 3)$ and $\psi_2(D_2) = (3; 0)$ (we assume that the state of $A$ corresponds to the state of $D_1$ and that $P(B|D_2, A, C)$ is specified s.t. the state of $C$ is revealed if the state of $A$, $B$ and $D_2$ are the same, and no knowledge is gained on $C$ if this is not the case).

These functions define two realizations of $\mathcal{I}$ who only differ on $\psi$, and when evaluating $\mathcal{I}$ w.r.t. these realizations we obtain two different optimal strategies for $D_1$ i.e. $\delta_{D_1} = d_1$ if $\psi = \psi_1$ and $\delta_{D_1} = d_2$ if $\psi = \psi_2$. From these strategies it can be seen that the utility function $\psi$ influences the optimal strategy for $D_1$ i.e. $\psi$ is relevant for $D_1$.  □

The example above can be seen as an instance of Metarule 2. Assume an arbitrary ID $\mathcal{I}$ and a realization of $\mathcal{I}$ in which the conditional probability function for any intermediate variable in a converging connection is as specified in the example above. From this structure we may deduce that, if $\psi'$ is relevant for a future decision $D'$ and $D$ is required for $D'$, then the utility function $\psi'$ is relevant for $D$; $D$ has an impact an $D'$.

**Example 4.** When deciding on $D_4$ in the ID depicted in figure 5, we seek to maximize $\psi'$. By the arguments given in the example above, it follows that both $\psi$ and $\psi'$ are relevant for $D_2$. Additionally, knowledge of $B$ may change our belief in $E$ when deciding on $D_4$, and since $B$ is influenced by $A$, which in turn is influenced by $D_1$, it follows that $D_1$ has an impact on $D_2$ ($A$ is required for $D_2$). Thus, both $\psi$ and $\psi'$ are relevant for $D_1$.

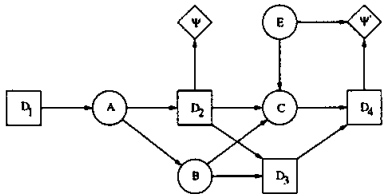

Figure 5: The figure represents a ID, in which $D_1$ may influence the chance variable $A$ required for $D_2$, indicating that both $\psi$ and $\psi'$ are relevant for $D_1$.

This can also be seen by assuming the two realizations consisting of the utility functions $\psi_1(D_2) = (0; 3)$ and $\psi_2(D_2) = (3; 0)$ together with functions corresponding to the ones specified in Example 2 and Example 3. When evaluating $\mathcal{I}$ w.r.t. these realizations we obtain two optimal strategies which differ on $D_1$ i.e. $\delta_{D_1} = d_1$ if $\psi = \psi_1$ and $\delta_{D_1} = d_2$ if $\psi = \psi_2$. From these strategies it can be seen that the utility function $\psi$ may influence the optimal strategy for $D_1$ i.e. $\psi$ is relevant for $D_1$.  □

The structural properties relating to the example above can, as for the previous example, be generalized to an arbitrary ID. Thus, based on the example above and the deductions made w.r.t. Example 3, we present the following rule.

**Rule 2.** Let $\mathcal{I}$ be a PID and let $<$ be an admissible total order for $\mathcal{I}$. The utility function $\psi$ is relevant for the decision variable $D$ w.r.t. $<$ if there exists a decision variable $D'$ s.t.

i) $D < D'$ and $\psi$ is relevant for $D'$ w.r.t. $<$.

ii) either

    a) $D$ is required for $D'$ w.r.t. $<$ or

    b) there exists a directed path in $\tilde{\mathcal{I}}$ from $D$ to a chance variable $X \in pred(D')$, and $X$ is required for $D'$ w.r.t. $<$.

This rule is equivalent to Metarule 2, since the rule covers all the cases where $D$ can have an impact on a future decision $D'$; $D$ has an impact on $D'$ if and only if $D$ is required for $D'$ or $D$ influences a variable required for $D'$.

Note that Rule 2 is not a complete structural rule as it refers to the term "required", which has not yet been characterized structurally. This is done in the following section.

### 3.3 REQUIRED VARIABLES - EXAMPLES AND RULES

Having established a method to identify the utility functions relevant for a decision variable, one might think that the required variables could be identified in the following way: Before constructing the moral graph, remove all utility functions not relevant for $D$ and then eliminate the variables as described in Section 3.1. However, the resulting neighbouring set $N(D)$ may still contain variables which are not required as can be seen from Figure 8: if we add the arc $(A, D)$ and remove the arc $(D''', \psi)$ then $A$ is not required for $D$ but $A \in N(D)$ when $D$ is eliminated.

A variable $X$ is required for a decision variable $D$ if $X$ is observed before $D$ and the state of $X$ may influence the optimal strategy for $D$. Since the optimal strategy for $D$ is dependent on the assumption that we always adhere to the maximum expected utility principle it follows that $X$ is required for $D$ if $X$ has an impact on $D$ or $X$ has an impact on a future decision variable $D'$, on which $D$ also has an impact. Hence, analogously to the metarules specifying the utility functions relevant for a decision variable, we present three metarules concerning the variables required for a given decision variable; according to the definition of a required variable we assume an admissible total order $<$ where a variable $X$ occurs before a decision variable $D$ under $<$.

**Metarule 4.** $X$ is required for $D$ if there is a realization s.t. when deciding on $D$ the state of $X$ has an impact on the expected utility for a utility function $\psi$ relevant for $D$ w.r.t. $<$.

**Metarule 5.** $X$ is required for $D$ if there is a realization and a future decision $D'$ s.t. $X$ has an impact on $D'$, and there exists a utility function $\psi$ relevant for both $D$ and $D'$ w.r.t. $<$.

**Metarule 6.** If none of the metarules above can be applied then $X$ is not required for $D$.

The following examples present a set of PIDs in which some of the required variables are identified. The properties described by these examples will be generalized to arbitrary PIDs, and they will serve as a basis for a theorem describing the structural constraints necessary and sufficient for a variable to be required for a given decision variable. In the examples we assume that chance variables, with no immediate predecessors, are given an even prior distribution.

**Example 5.** Consider the PID $\mathcal{I}$ depicted in figure 6. The utility function $\psi$ is relevant for $D$ for any admissible total ordering of $\mathcal{I}$, and since $\psi$ is functionally dependent on the chance variable $A$ it follows that $A$ is required for $D$; as $A$ is incompatible with $D$ there is an admissible total order $<$ with $A < D$.

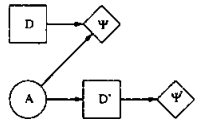

Figure 6: The figure represents a PID in which the chance variable $A$ is required for $D$.

□

The example above can be generalized to an arbitrary PID $\mathcal{I}$, assuming that $\mathcal{I}$ contains a decision variable $D$ and a variable $X$. If there exists an admissible total order $<$ s.t. $X$ occurs before $D$ under $<$ and $X$ is d-connected to a utility function $\psi$, relevant for $D$ w.r.t. $<$, given pred($D$), then $X$ is required for $D$; we can specify a realization of $\mathcal{I}$ s.t. the state of $X$ has an impact on the expected utility for $\psi$.

**Rule 3.** *Let $\mathcal{I}$ be a PID and let $D$ be a decision variable in $\mathcal{I}$. The variable $X$ is required for $D$ if there exists a utility function $\psi$ relevant for $D$ w.r.t. an admissible total order $<$ s.t. $X$ occurs before $D$ under $<$ and $X$ is d-connected to $\psi$ given pred($D$).*

This rule is equivalent to Metarule 4 as can be seen by expressing the metarule mathematically: $X$ is required for $D$ if $P(\text{dom}(\psi)|D, \text{pred}(D))$ is a function of $X$, and $\psi$ is a utility function relevant for $D$. The probability function $P(\text{dom}(\psi)|D, \text{pred}(D))$ is a function of $X$ if and only if $X$ is d-connected to $\psi$ given pred($D$).

The following examples show, that in order to identify all the variables required for a decision variable $D$, it is in general not sufficient only to consider the variables which directly influence the decision made w.r.t. $D$ (see Metarule 5).

**Example 6.** In the PID $\mathcal{I}$ depicted in figure 7 the chance variable $A$ may be observed before $D$. Moreover, $A$ is required for $D''$ since $\psi$ is relevant for $D''$ and $A$ is d-connected to $\psi$ given pred($D''$). Now, since $\psi$ is relevant for $D$ also, it follows that if $D < D''$ then $A$ and $D$ may both have an impact on $D''$. Additionally, if $A$ is observed prior to $D$ then the state of $A$ can be taken into account when deciding on $D$ i.e. $A$ is required for $D$.

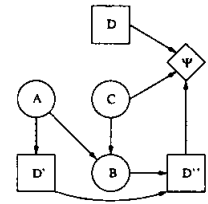

Figure 7: The figure represents a PID in which the variable $A$ is required for $D$.

This can also be seen by considering the utility function specified by Table 2, assuming that $P(B|A, C)$ has the properties of $P(C|D_2, B, E)$ specified in Example 3.

Table 2: The utility function $\psi(D, C, D'')$.

| $\psi(D,C,D'')$ | $C = c_1$ | $C = c_2$ |
|---|---|---|
| $D'' = d_1$ | (10; 2) | (1; 0) |
| $D'' = d_2$ | (0; 1) | (2; 10) |

The optimal strategy for $D$ relative to $A, D, D', B, D'', C$ is given by $\delta_D(a_1) = d_2$ and $\delta_D(a_2) = d_1$ which indicate, that $D$ is dependent on the state of $A$ ($A$ is required for $D$). □

In the example above, the conditional probability function for $B$ is specified s.t. the state of $C$ is revealed if the state of $A$ corresponds to the state of $B$, and no knowledge is gained on $C$ if the state of $A$ does not correspond to the state of $B$. Furthermore, the utility function $\psi$ relevant for both $D$ and $D''$ is specified s.t. the state of $C$ and the decision made w.r.t. $D$ influences $D''$. Thus, $A$ is required for $D''$ and therefore required for $D$ also.

Analogously to the previous examples, we may generalize this example by considering an arbitrary PID $\mathcal{I}$ and an admissible total order for $\mathcal{I}$, where a variable $X$ occurs before a decision variable $D$ and $X$ is required for a future decision $D'$, which has a relevant utility function in common with $D$. By specifying the realization of $\mathcal{I}$ according to the example above, it follows that the state of $X$ may influence the decision made w.r.t. $D'$. Moreover, since $D$ and $D'$ have a relevant utility function in common and $X$ is required for $D'$ we have that $X$ is required for $D$ also; the state of $X$ can be taken into account when deciding on $D$.

**Example 7.** Consider the PID $\mathcal{I}$ depicted in figure 8. Observing the chance variable $A$ may reveal the state of $X$, and when deciding on $D'''$ the observation of $X$ may change our belief in the state of $C$. Now, since $\psi$ is relevant for both $D$ and $D'''$ the decision made w.r.t. $D$ has an impact on $D'''$, and since the state of $A$ may influence the decision made w.r.t. $D'''$, and thereby the optimal strategy for $D$, it follows that the state of $A$ is relevant when deciding on $D$. That is, $A$ is required for $D$.

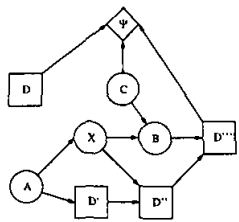

Figure 8: The figure represents a PID in which the chance variable $A$ is required for $D$.

This can also be seen by considering a realization of $\mathcal{I}$, where $P(X|A)$ is a deterministic function and $\psi(D, C, D''')$ and $P(B|X, C)$ correspond to $\psi(D, C, D'')$ and $P(A|D_1, B)$ respectively (see Example 3 and Table 2). When evaluating $\mathcal{I}$ w.r.t. $A, D, D', X, D'', B, D''', C$ the optimal strategy for $D$ is given by $\delta_D(a_1) = d_2$ and $\delta_D(a_2) = d_1$ i.e. $A$ is required for $D$.    □

In the example above, the state of $X$ is determined by the state of $A$, whereas the state of $C$ is determined by the state of $X$ and $B$ i.e. as for the previous examples the state of $C$ is revealed if the state of $X$ corresponds to the state of $B$, whereas no knowledge is gained on $C$ if the state of $X$ does not correspond to the state of $B$. Now, consider an arbitrary PID $\mathcal{I}$, and let $X$ denote a variable which occurs before the decision variable $D$ under $<$, and assume that $X$ is d-connected to a chance variable $Y \in \text{pred}(D')$ given $\text{pred}(D)$. If $Y$ is required for $D'(D < D')$ and $D'$ has a relevant utility function relevant in common with $D$ w.r.t. $<$, then we can specify a realization, as described in the example above, s.t. the optimal strategy for $D$ is dependent on the state of $X$. That is, $X$ is required for $D$.

**Rule 4.** *Let $\mathcal{I}$ be a PID and let $D$ be a decision variable in $\mathcal{I}$. Then the variable $X$ is required for $D$ if there exists a utility function $\psi$ relevant for $D$ w.r.t. an admissible total order $<$, where $X$ occurs before $D$ and:*

*i) there exists a decision variable $D'(D < D')$ s.t. $\psi$ is relevant for $D'$ w.r.t. $<$.*

*ii) $X$ is required for $D'$ or $X$ is d-connected to a chance variable $Y \in \text{pred}(D')$ given $\text{pred}(D)$ and $Y$ is required for $D'$.*

This rule is equivalent to Metarule 5, since the observation of $X$ can have an impact on a future decision $D'$ if and only if $X$ is required for $D'$ or $X$ influences a variable required for $D'$.

The rules 2 and 4 represent a set of simultaneous recursive structural constraints. The recursion terminates because it moves forward in the temporal ordering for each "call".

Based on the rules above we present the following theorem which defines the structural constraints necessary and sufficient for a variable to be required for a given decision variable.

**Theorem 3.** *Let $\mathcal{I}$ be an PID and let $D$ be a decision variable in $\mathcal{I}$. Then the variable $X$ is required for $D$ if and only if Rule 3 or Rule 4 (and Rule 1 and Rule 2) can be applied.*

*Proof.* The "if" part of the proof is apparent from the examples above. A mathematial proof of the "only if" part can be performed by closely following the elimination process when solving an ID. The basic idea is to postpone the calculations until a maximization is performed in order to calculate a strategy for a decision variable. That is, instead of marginalizing out a chance variable a script is produced, and when maximizing, the relevant scripts are identified. The details may be found in [Nielsen and Jensen, 1999].    □

Based on the previous rules we present the following rule characterizing the chance variables significant for a given decision variable; this rule is apparent from Rule 3 and Rule 4.

**Rule 5.** *Let $\mathcal{I}$ be an PID and let $D$ be a decision variable in $\mathcal{I}$ incompatibel with a chance variable $A$. Then $A$ is significant for $D$ if there exists a utility function $\psi$ relevant for $D$ w.r.t. an admissible total ordering $<$, where $A$ occurs immediately before $D$ s.t. :*

*i) $A$ is d-connected to $\psi$ given $\text{pred}(D)$ or*

*ii) there exists a decision variable $D'(D < D')$ s.t. $\psi$ is relevant for $D'$ and:*

   *a) $A$ is required for $D'$ or*

   *b) $A$ is d-connected to a chance variable $X \in \text{pred}(D')$ given $\text{pred}(D)$ and $X$ is required for $D'$.*

Additionally we have the folllowing corollaries as a consequence of theorem 3.

**Corollary 1.** *Let $\mathcal{I}$ be a PID and let $D$ be a decision variable in $\mathcal{I}$. Then the utility function $\psi$ is relevant for $D$ if and only if Rule 1 or Rule 2 can be applied.*

**Corollary 2.** *Let $\mathcal{I}$ be a PID and let $D$ be a decision variable in $\mathcal{I}$ incompatible with the chance variable $A$. Then $A$ is significant for $D$ if and only if Rule 5 can be applied.*

**Corollary 3.** *Let $\mathcal{I}$ be an ID and let $D$ be a decision variable in $\mathcal{I}$. Then $X$ is required for $D$ if and only if Rule 3 or Rule 4 can be applied.*

## 4  ALGORITHMS

In order to determine whether or not a PID defines a decision scenario, it is in principle necessary to investigate all admissible total orderings $<$ with a pair of incompatible variables being neighbours in $<$. However, the set of admissible orderings to investigate can be reduced substantially. E.g. if $(A, D)$ is an incompatible pair, then the ordering of the predecessors is of no importance. Also, if all pairs of incompatible successor nodes have been investigated and found commutatively irrelevant, then we can take any admissible order of the successors of $A$ and $D$. Thus, we start off with maximal pairs of incompatible pairs and work ourself backwards in the partial temporal ordering.

## 5  CONCLUSION

We have defined a PID as a generalization of the traditional ID by allowing a non-total ordering of the decision variables. Because the solution to a decision problem may be dependent on the temporal ordering of the decisions, we specified the class of PIDs whose solution is independent of the legal evaluation schemes i.e. the class of PIDs that represents a welldefined decision scenario. Additionally, we presented the constraints necessary and sufficient for a PID to be contained in this class.

The constraints were given in terms of the concept d-connectivity and are thus readable from the graphical structure. Based on these constraints an algorithm has been designed and implemented to determine whether or not a PID represents a welldefined decision scenario (the algorithm uses the methods in [Geiger et al., 1990] to determine d-connectivity). The algorithm has been tested on various PIDs, including those from the paper.